\documentclass[letterpaper, 10pt, conference]{ieeeconf}  

\IEEEoverridecommandlockouts                              
\usepackage{graphics} 
\usepackage{epsfig} 
\usepackage{lipsum}
\usepackage{mathptmx} 
\usepackage{times} 
\usepackage{amsmath} 
\usepackage{amssymb}  
\usepackage{acronym} 
\usepackage{booktabs}
\usepackage[style=base]{caption}
\usepackage{subcaption}

    \newacro{cnn}[CNN]{Convolutional Neural Network}
    \newacro{lstm}[LSTM]{Long Short-Term Memory}
    \newacro{bdlstm}[BDLSTM]{Bidirectional Long Short Term Memory}
    \newacro{bdgru}[BDGRU]{Bidirectional Gated Recurrent Unit}
    \newacro{seq2seq}[Seq2Seq]{sequence-to-sequence}
    \newacro{nlp}[NLP]{Natural Language Processing}
    \newacro{who}[WHO]{World Health Organization}
    \newacro{vru}[VRU]{Vulnerable Road User}
    \newacro{ap}[AP]{Average Precision}
    \newacro{gru}[GRU]{Gated Recurrent Unit}
    \newacro{rnn}[RNN]{Recurrent Neural Network}
    \newacro{bce}[BCE]{Binary Cross Entropy}

\title{\LARGE \bf
RNN-based Pedestrian Crossing Prediction using Activity and Pose-related Features
}

\author{J. Lorenzo$^{1}$, I. Parra$^{1}$, F. Wirth$^{2}$, C. Stiller$^{2}$, D. F. Llorca$^{1}$  and M. A. Sotelo$^{1}$
\thanks{
$^{1}$ Department of Computer Engineering, Universidad de Alcal\'a, Madrid, Spain
        {\{javier.lorenzod, ignacio.parra, david.fernandezl, miguel.sotelo\}@uah.es}\newline
$^{2}$ Institut f\"ur Mess- und Regelungstechnik, Karlsruher Institut f\"ur Technologie, Karlsruhe, Germany
{\{stiller, florian.wirth\}@kit.edu}
}
}

\begin{document}

\maketitle

\thispagestyle{empty}
\pagestyle{empty}

\begin{abstract}

Pedestrian crossing prediction is a crucial task for autonomous driving. Numerous studies show that an early estimation of the pedestrian's intention can decrease or even avoid a high percentage of accidents. In this paper, different variations of a deep learning system are proposed to attempt to solve this problem. The proposed models are composed of two parts: a CNN-based feature extractor and an RNN module. All the models were trained and tested on the JAAD dataset. The results obtained indicate that the choice of the features extraction method, the inclusion of additional variables such as pedestrian gaze direction and discrete orientation, and the chosen RNN type have a significant impact on the final performance.

\end{abstract}

\section{Introduction \& Related Work}
\label{sec:intro_related_work}
According to the \ac{who} \cite{who_report}, several efforts have been made in order to improve road safety, keeping the number of road deaths constant concerning the increase in both population and motorization.

The report additionally states that more than half of the reported fatalities are of \acp{vru}, with pedestrians and cyclists representing 26\% of all deaths. According to the EU \cite{EuropeanComission2018}, 37\% of road fatalities were in urban environments, and an additional two billion people are expected to be living in those areas by 2045, aggravating the problem \cite{who_report}.

Over the last decade, autonomous driving systems have evolved mainly due to the advent of Deep Learning. However, while tasks such as object classification and localization \cite{detsurvey} have been significantly developed and improved, the understanding of the environment continues to be a challenging problem. Referring to pedestrians, the ability to predict pedestrian crossing action in urban scenarios can help in the planning strategy, achieving a smoother and more human-like autonomous driving. Moreover, as it is explained in \cite{will-ped-cross-gavrila14}, an improvement in the anticipation time can lead to a considerable reduction of possible pedestrian injuries.

There are two main approaches related to pedestrian crossing prediction. The first one is the human motion-based approach. These methods try to infer pedestrian intention employing dynamics, whether using information extracted from 3D pose \cite{quintero18}, from image data and 3D position \cite{Rehder2017PedestrianPB} or using optical flow information \cite{stylesIV2019}. For a detailed overview of human motion methods, see the survey by Rudenko et al. \cite{motion_survey_gavrila2019}. Nonetheless, due to pedestrian complex dynamic behavior, position forecasting must be supported by additional information such as context-related. In \cite{kooij_context-based_2019}, authors based prediction on dynamic context variables such as distance to the car, and distance to the curb, in order to cope with sudden changes in dynamics. However, even if the prediction error decreases, the position forecasting does not take into account the majority of environmental variables which could affect the decision of the pedestrian. A higher-level approach, closer to the way the driver's mind works, tries to simplify the problem and, at the same time, capture information about the context and environment in a non-supervised way.\newline This second approach, based on action classification, tries to simplify the problem by getting closer to the driver's way of inferring pedestrian intention. One way to pursue this approach is by using \acp{cnn}. For example, in \cite{Rasouli_2017_ICCV} a pre-trained \ac{cnn} model is fine-tuned for this task. In the field of video action recognition, 3D \acp{cnn} have recently become also popular \cite{resnet3d}. As an example of the use of this architecture, in \cite{cnn3d-khaled}, a 3D \ac{cnn} spatio-temporal model is used together with an object detector and a tracking algorithm achieving real-time performance at $20$ fps.

\begin{figure}[t]
    \vspace{1em}
    \centering
    \includegraphics[width=\columnwidth]{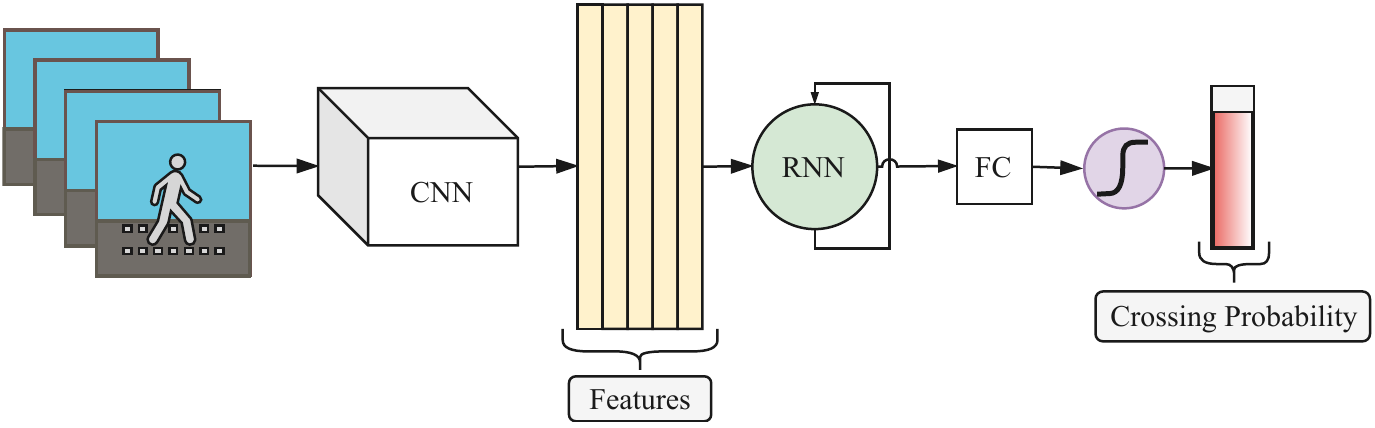}
    \caption{Diagram describing the proposed method.}
    \label{fig:general_seq2seq}
\end{figure}

In this paper, we propose a method for prediction of pedestrian crossing intention in one or more timesteps in the future, using data extracted from color videos recorded from inside a vehicle. Furthermore, additional information related to orientation, looking/gaze direction, movement state and image coordinates is used and compared with the model based solely on video input to observe possible improvements. The rest of the article is organized as follows: section \ref{sec:sys_desc} describes the proposed algorithms and the different architectures used. Experimental setup, including dataset selection, preprocessing methodology and experiments description, is detailed in section \ref{sec:exp_set}. In section \ref{sec:results}, experimental results are presented and discussed. Conclusions and future works are described in section \ref{sec:conc_fut_work}.

\section{System Description}
\label{sec:sys_desc}
The significant development of Deep Learning during the last decade has propelled the use of several variants of neural networks. In this work, two of these variants have been used: \acp{cnn}, used to extract features from pedestrian image sequences and \acp{rnn}, utilized to extract temporal information from these features. The proposed deep learning systems try to answer the question ``Is the pedestrian going to cross the street?" by approaching it as a sequence binary classification problem where we try to infer the intention in a future time horizon given an input sequence. In the next subsections, both the proposed problem and the architecture of the developed models will be discussed.

\subsection{Problem description}
\label{subsec:problem-descr}

The purpose of the proposed Deep Learning system is to predict the crossing intention of pedestrians by using temporal information provided by image sequences and other categorical variables.

The input sequences are defined as a set of features $\mathbf{X}_t = \{ X_{t-N}, X_{t-N+1}, \ldots, X_{t}\}$, where $N$ is the number of past frames and $t$ the current frame. The output is defined as a binary label $Y_{t+M}$ where $t+M$ is the index of the frame to be predicted. Thus, each pedestrian track with length $P$ is divided in $S=P-N-M$ subsequences, with $t \in \mathbb{N}: t \in [ N, P-M-1 ]$. The remaining section will discuss the architecture followed by our model, explaining the role of each module separately.

\subsection{General model architecture}

\begin{figure}[t]
    \vspace{1em}
    \centering
    \includegraphics[width=\columnwidth]{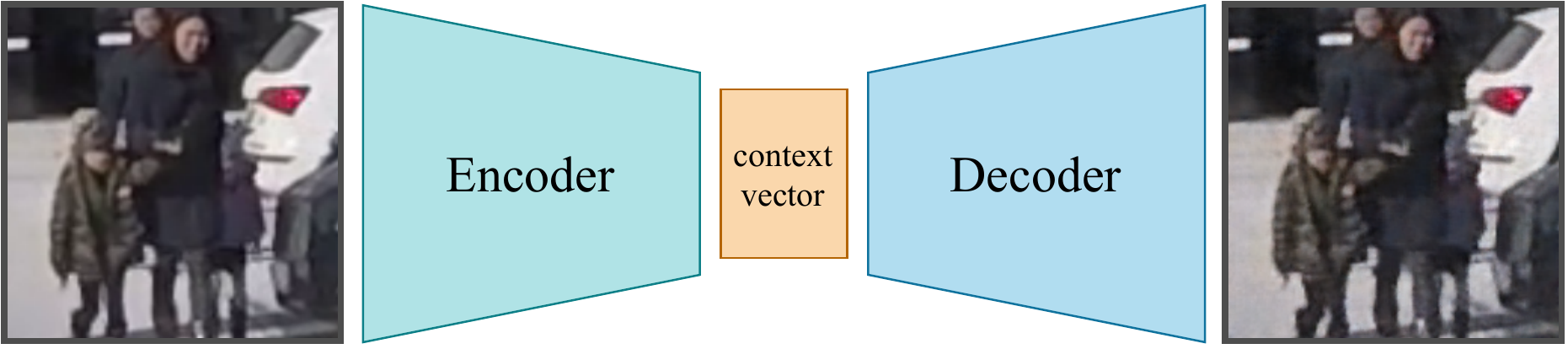}
    \caption{High-level diagram describing a Convolutional Autoencoder. Output image example corresponds to SegNet-based autoencoder.}
    \label{fig:conv-autoencoder}
\end{figure}

The proposed system is composed by two main modules: a \textbf{feature extractor}, used to get useful information from image data and a \textbf{many-to-one \ac{rnn} module}. At a high level, features extracted are introduced to the \ac{rnn} module. Output of the \ac{rnn} module is introduced in a fully-connected layer, and its output is passed through a sigmoid in order to get the predicted probability of crossing action in the trained time horizon. This architecture is represented in Fig. \ref{fig:general_seq2seq}.
\subsection{Feature extraction}
\label{subsec:feature-extraction}
Input features are extracted from color video sequences using three alternative techniques:

\begin{itemize}
    \item Pretrained \ac{cnn} models from ResNet family \cite{resnet15} and from ResNeXt family \cite{resnext16}. All models are pretrained on ImageNet \cite{imagenet_cvpr09}). The network was modified by cutting off the last fully connected layer and obtaining the features from the average pool layer output.
    \item Convolutional autoencoder with previous pre-trained ResNet34 used as encoder \cite{dlbook}. An autoencoder is a type of encoder-decoder variant which is trained for the task of input reconstruction in a self-supervised manner. After the training process, the encoder is separated from the decoder and used as a feature extraction method. A high-level diagram of this architecture is shown in Fig. \ref{fig:conv-autoencoder}. The network was trained with a learning rate of $10^{-3}$ and using \ac{bce} loss. 
    \item SegNet-based autoencoder \cite{segnet15}. This method is pre-trained on Cars Dataset \cite{car-dataset13} and obtained from \cite{autoencoder_2020}.
\end{itemize}

No fine-tuning has been applied to any of the feature extractors during the \ac{rnn} models training.

Following the same pooling strategy as in the pretrained ResNet34, output features of both encoders extracted from trained autoencoders, with size $512\times 7 \times 7$, are averaged with a pooling layer with a $7\times 7$ kernel, obtaining a $512\times 1\times 1$ tensor. The obtained tensor is flattened in order to obtain a one-dimensional vector of size $512$.

In some experiments, categorical variables are used as inputs along with images. These variables are embedded in order to learn their multidimensional relationship between their categories. These embeddings are learned during training, and their dimension for each category is established following the heuristic proposed in the course imparted by Jeremy Howard \cite{fastaicourse}: $min(Int(N_c / 2 + 1), 50)$ where $N_c$ is the number of categories of the variable (cardinality).

\subsection{\ac{rnn} module}
\label{subsec:enc_dec_mod}
For the recurrent module of the system, two variants of \acp{rnn} are used: \ac{lstm} \cite{hochreiter1997long} and \ac{gru} \cite{gru14}. These variants help in the fight of \acp{rnn} vanishing gradient problem. The main difference between \acp{gru} and \acp{lstm} is that \acp{gru} are computationally more efficient and according to \cite{grulstmcomparison14}, they achieve similar results in sequence modeling problems.

Bidirectional variants of \acp{lstm} and \acp{gru} are also used on the experiments in order to test if the additional information of the reversed sequence can improve the understanding of the problem.
\section{Experimental Setup}
\label{sec:exp_set}
In the following section, all experiments carried out will be detailed separately. Unless otherwise noted, 
\ac{lstm} module with the pre-trained ResNet50 used for feature extraction is the selected choice for the tests.

\subsection{Image Data Preparation}

\begin{figure*}[t]
\vspace{2em}
    \centering
    \includegraphics[keepaspectratio=true, width=\textwidth]{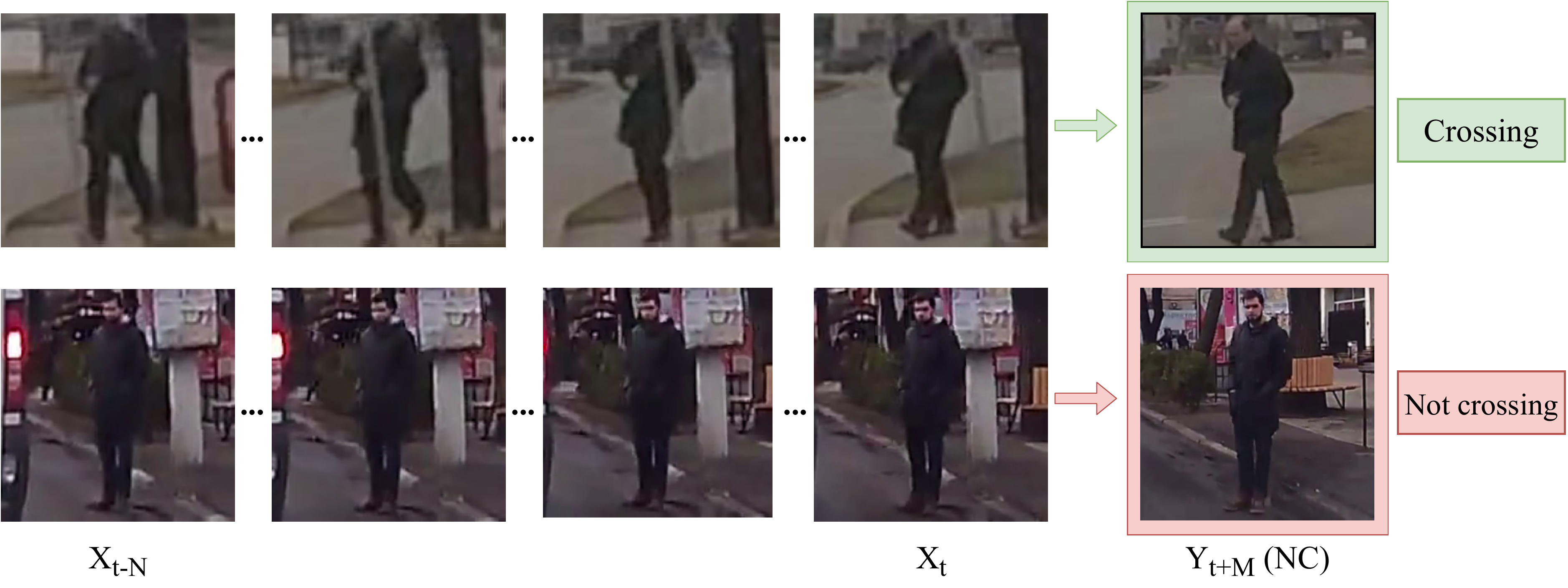}
    \caption{Two examples of JAAD pedestrian sequences: a crossing (top) and not crossing (down) situation.}
    \label{fig:my_label}
\end{figure*}

All models have been trained on the JAAD \cite{jaad_dataset}, a naturalistic dataset focused on the behavior of pedestrians during their road-crossing action. It comprises 346 videos filmed inside a vehicle of duration ranging from 5 to 10 seconds. Their format varies both in frame rate and in resolution. There are 8 videos at 60 fps and 10 videos in HD resolution ($\mathrm{1280} \times \mathrm{720}$). The rest of the videos are filmed at 30 fps in FHD resolution ($\mathrm{1920} \times \mathrm{1080}$)

Default split sets for training and testing suggested by the authors have been used in order to encourage possible future comparisons with other algorithms. In this split, HD videos are excluded, in addition to another set that presents low visibility (night scenes, heavy rain), totaling 323 videos. The videos at $60$ fps included in these splits have been lowered to $30$ fps.

The input to the model is composed of image sequences and, in some variants, categorical variables. Image sequences are extracted using the ground truth 2D bounding box annotations of pedestrians with crossing behavior. The height and width of them are equalized in order to avoid image deformation. All sequences are filtered by occlusion level and the bounding box height. Fully occluded samples and bounding boxes with height lower than $50$ pixels have been removed only in the training set, leaving all the other sets unchanged, in order to see the behavior of the model in challenging situations.

Finally, in order to meet the input restrictions of feature extraction methods, images are resized to $\mathrm{224} \times \mathrm{224}$ (size used in training) and standardized using the per-channel mean and deviation of ImageNet.

\subsection{Feature extraction method importance}

Various tests were performed changing the feature selection method, one for each option on the list in subsection \ref{subsec:feature-extraction}. The reason for using autoencoders is to test if features extracted with a method specialized on the reconstruction of images help the network in its training process more than a classification pre-trained network.

\subsection{Rescaling image features and normalization}

Output data of the average pooling layer have a range between $0$ and a maximum value which depends on the input image and also on the feature extractor. We tried a rescaling approach to test if there is any improvement in the results. Rescaling is performed by dividing the sequence of image features between the maximum value in the batch, obtaining data between $0$ and $1$.

\subsection{Influence of additional variables}

Three categorical variables related to pedestrians and extracted from ground truth annotations have been used to study their influence on predictions: looking/gaze direction, orientation and state of movement. Looking direction is a binary variable, whose value is $1$ if the pedestrian looks at the vehicle and $0$ otherwise. The orientation variable has the following categories and are defined relative to the car: \textit{front} ($0$), \textit{back} ($1$), \textit{left} ($2$) and \textit{right} ($3$). State of movement has two possible values: \textit{standing} and \textit{moving}. Another variable used is the bounding box center $(u_{c}, v_{c})$, extracted from groundtruth annotations and divided by the maximum of each dimension in order to achieve independence from the camera sensor used. 

The output of each embedding layer and the center of the bounding box are concatenated to the feature vector. As a result, the input vector used in the \ac{rnn} module increases its size from $3$ to $9$.

\subsection{\ac{lstm} versus \ac{gru}}

As mentioned in subsection \ref{subsec:enc_dec_mod}, we perform an study on the influence of the type of \ac{rnn} chosen, and their bidirectional variants. With this objective in mind, four \ac{rnn} models are compared with the same hyperparameter configuration: \ac{lstm}, \ac{gru}, \ac{bdlstm} and \ac{bdgru}.

\subsection{Hyperparameter search}

After an ablation study using grid search, the configuration used for the model is the following:
\begin{itemize}
    \item \ac{rnn} hidden dimension: 4
    \item Number of stacked \ac{rnn} layers: 1
    \item Dropout (applied to \ac{rnn} output): 0.5
\end{itemize}
The simplicity of the network is due to the trend towards overfitting of more complex networks.

\subsection{Training configuration}

PyTorch \cite{pytorch} has been the framework chosen to carry out the experiments. All experiments have been trained and tested on a single NVIDIA GTX TITAN X GPU. We have used Adam \cite{kingma2014adam} as an optimizer with a learning rate of $10^{-4}$. The loss function used for training is the \ac{bce} loss. To make computations deterministic, a fixed random seed has been established in all pseudorandom number generators. Finally, to avoid unnecessary processing, validation patience with a value of five has been set, i.e., if validation losses stop improving during five epochs, the training will end.

\section{Results}
\label{sec:results}
The metrics used to compare these results are \textit{accuracy}, \textit{precision}, \textit{recall} and, finally \ac{ap} score, calculated as a weighted sum, following equation \ref{eq:ap}, where $R$ is recall, $P$ is precision and $n$ refers to the threshold number. All metric values are percentages.
\begin{equation}
\label{eq:ap}
    \text{AP} = \sum_n (R_n - R_{n-1}) P_n
\end{equation}

\subsection{Feature extraction method Importance}

\begin{table}[ht]
    \centering
    \caption{Different feature extraction methods results}
    \label{tbl:feat}
    \begin{tabular}{lllll}
        \toprule
        \textbf{Method} &
        {Acc.} &
        {P} &
        {R} &
        {AP}\\
        \midrule
        ResNet18  & 62.68 & 62.87 & 98.82 & 69.08\\
        ResNet34  & 63.32 & 65.71 & 86.75 & 74.33\\
        ResNet50  & 65.75 & 69.63 & 80.43 & 75.62\\
        ResNet101  & 68.95 & 71.29 & 84.49 & 77.16\\
        ResNet152  & 62.53 & 64.26 & 90.59 & 75.85\\
        ResNeXt50  & \textbf{70.04} & \textbf{74.96} & 78.39 & 79.87\\
        ResNeXt101  & 69.45 & 74.14 & 78.73 & \textbf{81.20}\\
        ConvAE-ResNet  & 62.67 & 62.67 & \textbf{100.00} & 61.64\\
        ConvAE-SegNet & 62.67 & 62.67 & \textbf{100.00} & 68.46\\
        \bottomrule
    \end{tabular}
\end{table}

Results per method are given in table \ref{tbl:feat}, pre-trained models from ResNet and ResNeXt families, obtain better results than self-trained ones. The increase in the complexity of the network is directly related to the increase in all performance metrics. One possible reason is the difference in training data size and diversity between ImageNet \cite{imagenet_cvpr09} and JAAD \cite{jaad_dataset} or CARS dataset \cite{car-dataset13}. 

Although the images are reconstructed quite accurately in the self-trained extractors, the output features of the encoder lack useful information for the \ac{rnn} module. This is shown in the recall value of 100\%, which means that the model has converged in predicting that every pedestrian will cross. This problem may be caused by the use of an average pooling layer after training since, in pre-trained models, average pooling is used during the training stage.

\subsection{Rescaling image features and normalization}

Rescaling input image features contributes to an improvement in the results (see table \ref{tbl:rescaling}). These results show that the high variation in input features penalized the learning process.

\begin{table}[ht]
    \centering
    \caption{Rescaling image features results}
    \label{tbl:rescaling}
    \begin{tabular}{lllll}
        \toprule
        \textbf{Normalization type} &
        {Acc.} &
        {P} &
        {R} &
        {AP}\\
        \midrule
        None                & 65.75 & 69.63 & 80.43 & 75.62\\
        Rescaling           & 65.89 & 70.75 & 77.70 & 76.84\\
        \bottomrule
    \end{tabular}
\end{table}

\subsection{Influence of additional variables}

The incorporation of all additional variables improves the \ac{ap} from $75.62$\% to a $80.00$\% (see table \ref{tbl:additional-variables}). This result shows that the incorporation of meaningful data can act as a regulatory factor to allow greater learning generalization. Individually, orientation and looking direction are the variables with more weight followed by the state of movement. Those variables are also used by drivers when they infer the pedestrians' crossing intentions (e.g. a pedestrian walking towards the road and a pedestrian at the curb looking at the driver's car are more likely to cross than a pedestrian walking parallel to the car and suddenly stopping). In the case of the bounding box center in the image, it has less weight. This is probably due to its relativeness and high variation as it belongs to the image coordinate system.

\begin{table}[ht]
    \centering
    \caption{Influence of additional variables results}
    \label{tbl:additional-variables}
    \begin{tabular}{lllll}
        \toprule
        \textbf{Variable} &
        {Acc.} &
        {P} &
        {R} &
        {AP}\\
        \midrule
        None            & 65.75 & 69.63 & 80.43 & 75.62\\
        Looking         & 65.13 & 67.28 & 86.37 & 76.94\\
        Orientation     & 67.12 & 69.96 & 83.30 & 77.71\\
        Bbox center     & 64.78 & 66.59 & 87.91 & 76.15\\
        Movement        & 68.76 & 72.71 & 80.30 & 76.75\\
        All             & 68.82 & 74.20 & 77.03 & 80.00\\
        \bottomrule
    \end{tabular}
\end{table}

\subsection{\ac{lstm} versus \ac{gru}}

\begin{table}[ht]
    \centering
    \caption{RNN selection results}
    \label{tbl:grub}
    \begin{tabular}{lllll}
        \toprule
        \textbf{Method} &
        {Acc.} &
        {P} &
        {R} &
        {AP}\\
        \midrule
        \ac{lstm}   & 65.75 & 69.53 & 80.43 & 75.62\\
        \ac{gru}    & 62.65 & 62.84 & 98.88 & 64.25\\
        \ac{bdlstm} & 67.33 & 69.20 & 86.28 & 79.07\\
        \ac{bdgru}  & 67.62 & 76.00 & 70.66 & 80.19\\
        \bottomrule
    \end{tabular}
\end{table}

According to table \ref{tbl:grub}, in this problem, additional temporal information provided by bidirectionality can improve the results of an \ac{lstm}-based network. \ac{gru} obtains worse results than \ac{lstm} and in the case of the bidirectional variants, both \acp{rnn} improve the results, with \ac{bdgru} performing slightly better than the \ac{bdlstm}. This can be due to the high dropout used and the fixed seed used for reproducibility.

\subsection{Final results}
\begin{table}[ht]
\vspace{1em}
    \centering
    \caption{Best model for each improvement configuration}
    \label{tbl:best-model}
    \begin{tabular}{ccccc}
        \toprule
        {all add. var.}     &
        {reescaling}        &
        {best \ac{rnn}}     &
        {best feat. extr.}  &
        {AP} \\
        \midrule
        -          & -           & -            & -            & 75.62 \\
        \checkmark  & -             & -             & -            & 80.00 \\
        -          & \checkmark    & -             & -              & 76.84 \\
        -           & -             & \checkmark    & -              & 80.19 \\
        -           & -             &     & \checkmark              & 81.20 \\
        \checkmark  & \checkmark    & \checkmark    & \checkmark    & 83.34 \\
        \bottomrule
    \end{tabular}
\end{table}

To see the effect of the above experiments together, a model has been trained including all of the previous upgrades. According to \ac{ap} scores in table \ref{tbl:best-model}, improvements work well together with an increase of more than $8$\% concerning the simpler model. 

\begin{figure*}[ht]
    \centering
    \includegraphics[keepaspectratio=true, width=\textwidth]{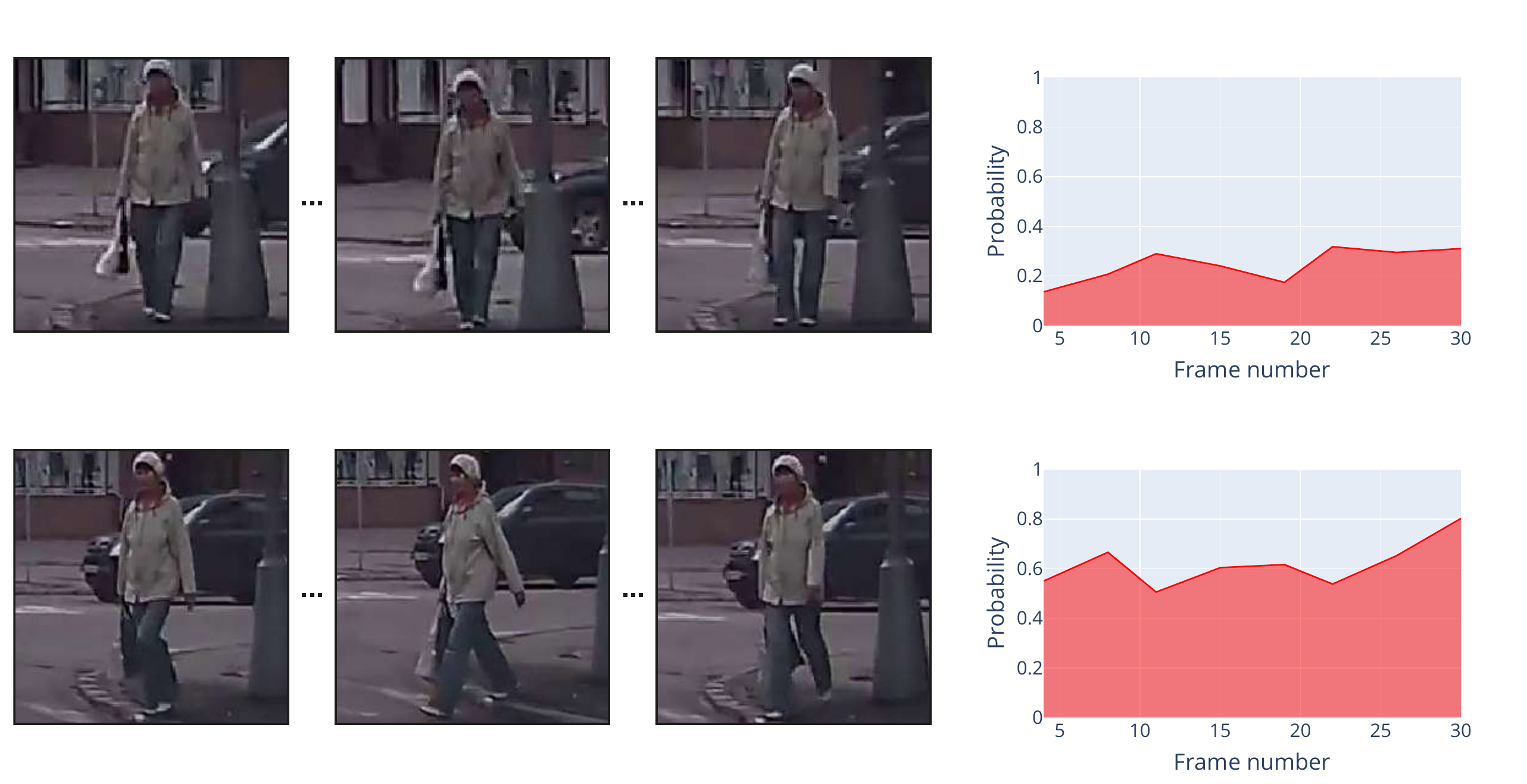}
    \caption{Two examples in test set. The top one represents a non-crossing sequence and at the bottom, a crossing one. Left graphics show the output crossing probability at eight future time steps between 0 and 1 seconds (0 and 30 frames).}
    \label{fig:qual}
\end{figure*}

\subsection{Qualitative results}

In figure \ref{fig:qual}, two example sequences are shown with the input image sequence at the left and the output crossing probability at the right. The model used in this experiment is the best model from table \ref{tbl:best-model} with a change in the output dimension. Instead of outputting the crossing probability one second in the future, the output is split into eight equispaced time steps between $0$ and $1$ second. Both sequences belong to the same pedestrian. In the top one, the pedestrian is not going to cross in one second in the future, and in the bottom one, the pedestrian is beginning to cross. As the graphs show, the probability of crossing is low in the first time step of the top graph, but this value is doubled at the end of the prediction, indicating a possible future crossing, which becomes more likely in the bottom case.

\subsection{Dataset limitations}

\begin{figure}[ht]
\vspace{1em}
     \centering
     \begin{subfigure}[b]{0.49\columnwidth}
         \centering
         \includegraphics[width=0.95\textwidth]{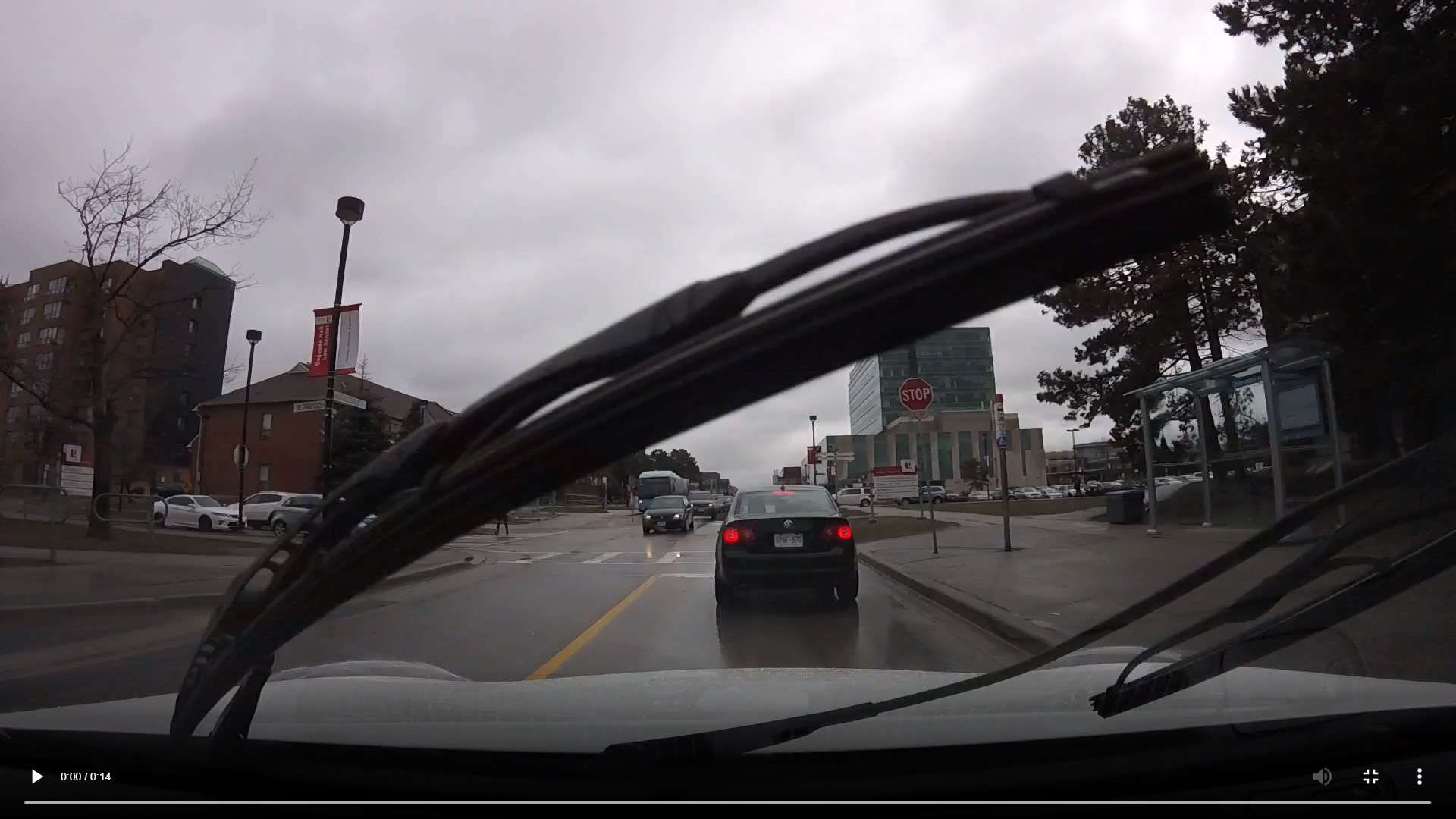}
         \caption{Wipers occlusion}
         \label{fig:limpia}
     \end{subfigure}
     \hfill
     \begin{subfigure}[b]{0.49\columnwidth}
         \centering
         \includegraphics[width=0.95\textwidth]{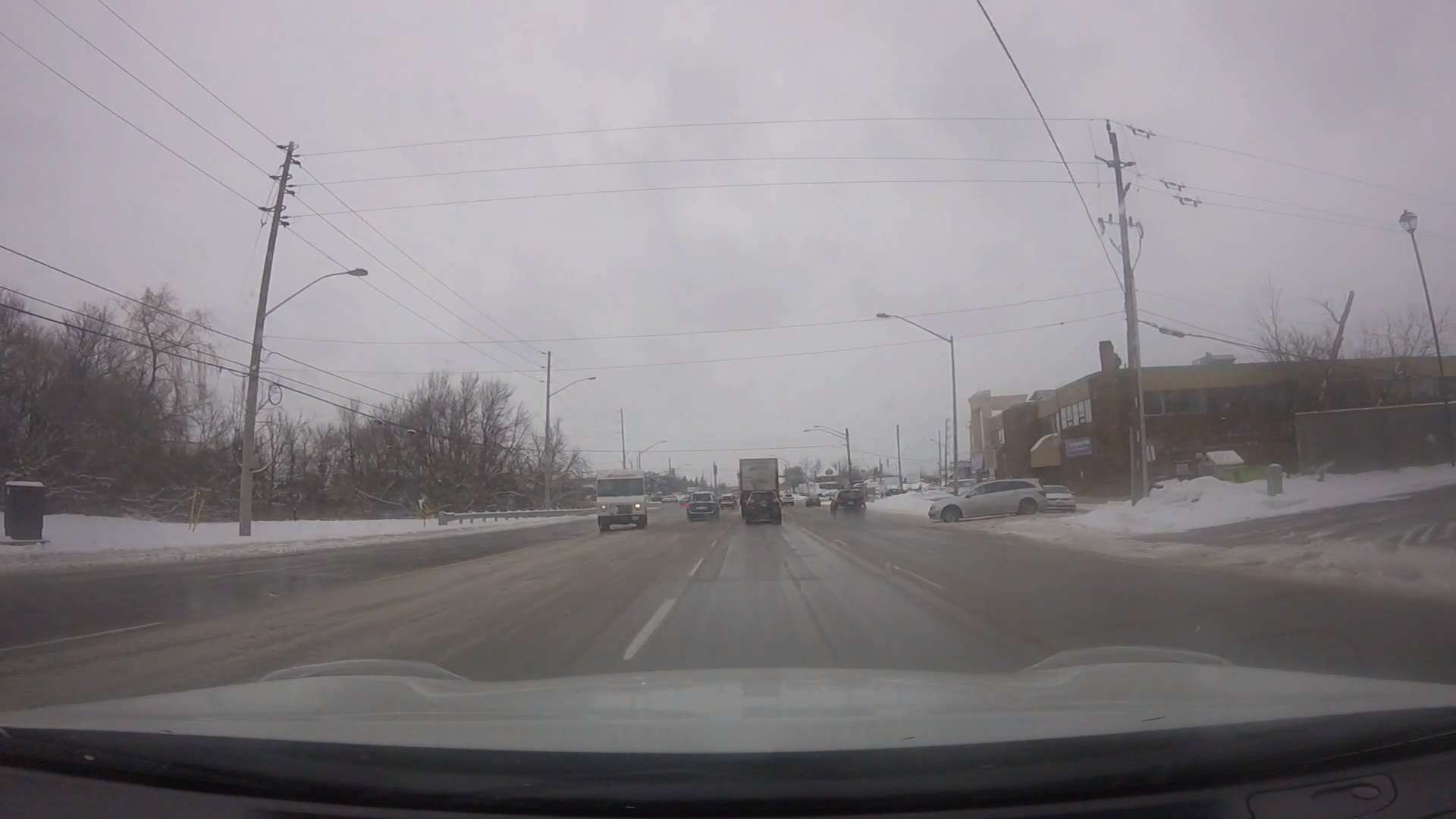}
         \caption{Bad weather conditions}
         \label{fig:nieve}
     \end{subfigure}
     
     \begin{subfigure}[b]{0.49\columnwidth}
         \centering
         \includegraphics[width=0.95\textwidth]{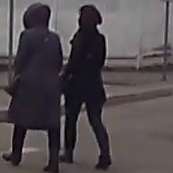}
         \caption{Non relevant pedestrians}
         \label{fig:norelev}
     \end{subfigure}
     \hfill
         \begin{subfigure}[b]{0.49\columnwidth}
         \centering
         \includegraphics[width=0.95\textwidth]{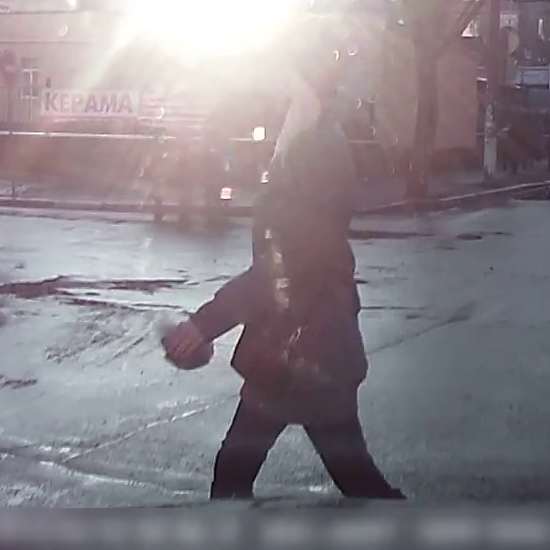}
         \caption{Reflections}
         \label{fig:refl}
     \end{subfigure}
        \caption{Some defiant cases from JAAD dataset}
        \label{fig:defiant}
\end{figure}

JAAD dataset is one of the few datasets focused on pedestrian behavior. However, it is composed of short videos. Besides there are challenging situations that affect training: windshield wipers occlusion, bad weather conditions (raining, snowing) and reflections on the windshield. Additionally, small pedestrians are a problem that can be filtered easily, but this is not the case for non-relevant pedestrians i.e., pedestrians who are crossing or not but are not in the path of the vehicle. Filtering out these problems can lead to better training convergence, but at the same time, it leads to a loss of training data. In Fig. \ref{fig:defiant} some examples of those challenging situations discussed before are shown. 

\section{Conclusions \& Future Work}
\label{sec:conc_fut_work}
A method based on a \ac{cnn} feature extractor and a \ac{rnn} many-to-one module has been proposed to predict pedestrian crossing action in the future. Image and categorical data have been the chosen sources of information for the model. Experiments carried out have shown that pre-trained networks can provide better temporal information than autoencoders. The inclusion of additional data can improve the results, as well as the use of bidirectional \ac{lstm}. Applying all improvements at the same time rises \ac{ap} score more than $8$ \%. These results are encouraging, and they show a way ahead in the development of more reliable and secure intention prediction systems.
\newline
As stated in the discussion, the JAAD dataset is useful for tasks such as detection and tracking, but not for video understanding.
For this reason, the PIE dataset \cite{Rasouli2019PIE} will be considered in future work in order to develop models. New context and local variables could also be studied, such as 3D pose, kinematics, relative distances and presence of traffic lights or zebra crossings. Concerning hyperparameter optimization, non-exhaustive search methods (e.g., Bayesian Optimization methods) could be applied to the training process. Finally, different strategies can be followed by the feature extractor output instead of averaging all output channels in a single vector (e.g., attention mechanisms or the use of convolutional \acp{lstm} \cite{xingjian2015convolutional}).
\section{Acknowledgements}
This work was funded by Research Grants S2018/EMT-4362  (Community Reg. Madrid), DPI2017-90035-R  (Spanish Min. of Science and Innovation), BRAVE Project, H2020, Contract \#723021 and by Universidad de Alcalá, via a predoctoral grant to the first author (FPI-UAH). It has also received funding from the Electronic Component Systems for European Leadership Joint Undertaking under grant agreement No 737469 (AutoDrive Project). This Joint Undertaking receives support from the European Unions Horizon 2020 research and innovation programme and Germany, Austria, Spain, Italy, Latvia, Belgium, Netherlands, Sweden, Finland, Lithuania, Czech Republic, Romania, Norway.
\bibliographystyle{IEEEtran}
\bibliography{refs}

\end{document}